\documentclass[runningheads,a4paper]{llncs}

\usepackage{amssymb}
\usepackage{graphicx}

\usepackage{url}
\urldef{\mailsa}\path|luo@ms.k.u-tokyo.ac.jp|    
\newcommand{\keywords}[1]{\par\addvspace\baselineskip
\noindent\keywordname\enspace\ignorespaces#1}

\begin{document}

\mainmatter  

\title{Misdirected Registration Uncertainty}

\titlerunning{\scriptsize{Misdirected Registration Uncertainty}}

%
%
\author{Jie Luo$^{1,2}$, Karteek Popuri$^3$,  Dana Cobzas$^4$, Hongyi Ding$^5$ \\ William M. Wells III$^{2,6}$ and Masashi Sugiyama$^{7,1}$}
\authorrunning{}

	
\institute{$^1$Graduate School of Frontier Sciences, The University of Tokyo, Japan\\ $^2$Radiology Department, Brigham and Women's Hospital, Harvard Medical School, USA\\ $^3$School of Engineering Science, Simon Fraser University, Canada\\ $^4$Computing Science Department, University of Alberta, Canada \\ $^5$Department of Computer Science, The University of Tokyo, Japan \\ $^6$ Computer Science and Artificial Intelligence Laboratory, Massachusetts Institute of Technology, USA \\ $^7$ RIKEN Center for Advanced Intelligence Project, Japan \\
	\mailsa 
}

%
%

\toctitle{Lecture Notes in Computer Science}
\tocauthor{Authors' Instructions}
\maketitle

\vspace{-5mm}

\begin{abstract}

\emph{Being a task of establishing spatial correspondences, medical image registration is often formalized as finding the optimal transformation that best aligns two images. Since the transformation is such an essential component of registration, most existing researches conventionally quantify the registration uncertainty, which is the confidence in the estimated spatial correspondences, by the transformation uncertainty. In this paper, we give concrete examples and reveal that using the transformation uncertainty to quantify the registration uncertainty is inappropriate and sometimes misleading. Based on this finding, we also raise attention to an important yet subtle aspect of probabilistic image registration, that is whether it is reasonable to determine the correspondence of a registered voxel solely by the mode of its transformation distribution.}

\keywords{Image registration, Uncertainty}
\end{abstract}

\section{Introduction}

Medical image registration is a process of establishing anatomical or functional correspondences between images. It is often formalized as finding the optimal transformation that best aligns two images \cite{Sotiras}. Since many important clinical decisions or analysis are based on registered images, it would be useful to quantify the intrinsic uncertainty, which is a measure of confidence in solutions, when interpreting the image registration results.

Among all methods that characterize the uncertainty of non-rigid image registration, the most mainstream, or perhaps the most successful framework is probabilistic image registration (PIR) \cite{Cobzas,Simpson,Risholm,Lotfi,Popuri,Wassermann,Simpson2,Heinrich}. Unlike point-estimate registration methods that report a unique set of transformation parameters, PIR models the transformation parameters as a random variable and estimates a distribution over them. PIR methods can be broadly categorized into discrete probabilistic registration (DPR) and continuous probabilistic registration (CPR). The transformation distribution estimated by DPR and CPR have different forms. DPR discretizes the transformation space into a set of displacement vectors. Then it uses discrete optimization techniques to compute a categorical distribution as the transformation distribution \cite{Cobzas,Lotfi,Popuri,Heinrich}. CPR is essentially a Bayesian registration framework, with the estimated transformation given by a multivariate continuous posterior distribution \cite{Simpson,Risholm,Wassermann,Simpson2}. A remarkable advantage of PIR is that its registration uncertainty can be naturally obtained from the distribution of transformation parameters, and further utilized to benefit the subsequent clinical tasks\cite{Risholm,Risholm2,Simpson3}.

\subsubsection{Related Work}
Image registration refers to the process of finding spatial correspondences, hence the uncertainty of registration should be a measure of the confidence in spatial correspondences. However, since the transformation is such an essential component of registration, in the PIR literature, most existing works do not differentiate the transformation uncertainty from the registration uncertainty. Indeed, the conventional way to quantify the registration uncertainty is to employ summary statistics of the transformation distribution. Applications of various summary statistics have been found in previous researches: the Shannon entropy and its variants of the categorical transformation distribution were used to measure the registration uncertainty of DPR \cite{Lotfi}. Meanwhile, the variance \cite{Simpson}, standard deviation \cite{Simpson2}, inter-quartile range \cite{Risholm} and covariance Frobenius norm \cite{Wassermann} of the transformation distribution were used to quantify the registration uncertainty of CPR. In order to visually assess the registration uncertainty, each of these summary statistics was either mapped to a color scheme, or an object overlaid on the registered image. By inspecting the color of voxels or the object's geometry, clinicians can infer the registration uncertainty, which suggests the confidence they can place in the registered image.

It is acknowledged that registration uncertainty should be factored into clinical decision making. This work mainly investigates whether those summary statistics of the transformation distribution truly give insight into the registration uncertainty. If clinicians are misdirected from the registration uncertainty to the transformation uncertainty, and hence be conveyed by the false amount of uncertainty with respect to the established correspondence, it can cause detrimental effects on their performance.

In the following sections, we use concrete examples and reveal that using the transformation uncertainty to quantify the registration uncertainty is inappropriate and sometimes misleading. Based on this finding, we also raise attention to an important yet subtle aspect of PIR, that is whether it is reasonable to determine the correspondence of a registered voxel solely by the mode of its transformation distribution.

\section{Misdirected Registration Uncertainty}

Most existing works do not differentiate the transformation uncertainty from the registration uncertainty. In this section, we give concrete examples and further point out that it is inappropriate to quantify the registration uncertainty by the transformation uncertainty. For the convenience of illustration, we use Random Walker Image Registration (RWIR) method as the PIR scheme in all examples \cite{Cobzas,Lotfi,Popuri}. 

\subsection{The RWIR Set Up}

In the RWIR setting, let $I_\mathrm{f}$ and $I_\mathrm{m}$ respectively be the fixed and moving image $I_\mathrm{f}  ,I_\mathrm{m}: \Omega_I\rightarrow\mathbb{R},\Omega_I\subset\mathbb{R}^d, d=2\: or \: 3$. RWIR discretizes the transformation space into a set of $K$ displacement vectors $\mathcal{D} = \{\mathbf{d}_k\}_{k=1}^K, \mathbf{d}_k\in\mathbb{R}^d$. These displacement vectors radiate from voxels on $I_\mathrm{f}$ and point to their candidate transformation locations on $I_\mathrm{m}$. The corresponding label for $\mathbf{d}_k$, which can be intensity values or tissue classes at those locations, are stored in  $\mathcal{I}=\{I(\mathbf{d_k})\}_{k=1}^K$. For every voxel $v_i$, the algorithm computes a unity-sum probabilistic vector $\mathcal{P}(v_i)=\{P_k(v_i)\}_{k=1}^K$ as the transformation distribution. $P_k(v_i)$ is the probability of displacement vector $\mathbf{d}_k$. In a standard RWIR, the algorithm takes a displacement vector that has the highest probability in $\mathcal{P}(v_i)$ as the most likely transformation $\mathbf{d}_m$. The corresponding label of $\mathbf{d}_m$ in $\mathcal{I}$ is assigned to voxel $v_i$ as its established correspondence.

Conventionally, the uncertainty of registered $v_i$ is quantified by the Shannon entropy of the transformation distribution $\mathcal{P}(v_i)$. Since RWIR takes $\mathbf{d}_m$ as its``point-estimate", the entropy provides a measure of how disperse the rest of displacement vectors in $\mathcal{D}$ are from $\mathbf{d}_m$. If other displacement vectors are all equally likely to occur as $\mathbf{d}_m$, then the entropy is maximal, because it is completely uncertain which displacement vector should be chosen as the most likely transformation. When the probability of $\mathbf{d}_m$ is much higher than the other displacement vectors, the entropy decreases, and it is more certain that $\mathbf{d}_m$ is the right choice. For example, assuming $\mathcal{P}(v_l)$ and $\mathcal{P}(v_r)$ are two discrete transformation distribution for voxels $v_l$ and $v_r$ respectively. As shown in Fig.1, $\mathcal{P}(v_l)$ is uniformly distributed, and its entropy is $E(\mathcal{P}(v_l))=2$. $\mathcal{P}(v_r)$ has an obvious peak, hence its entropy is $E(\mathcal{P}(v_r))\approx1.36$, which is lower than $E(\mathcal{P}(v_l)$.

\vspace{-4mm}

\begin{figure}
	\centering
	\includegraphics[height=2.3cm]{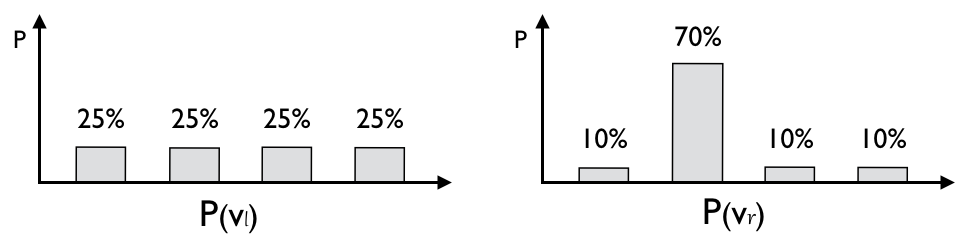}
	\vspace{-2mm}
	\caption{Discrete distribution $\mathcal{P}(v_l)$ and $\mathcal{P}(v_r)$.}
	\label{fig:correct1}
\end{figure}
\vspace{-8mm}

\subsection{Transformation Uncertainty and Registration Uncertainty}

For a registered voxel, the entropy of its transformation distribution is usually mapped to a color scheme. Clinicians can infer how uncertain the registration is by the color of that voxel. However, does the conventional uncertainty measure, which is the entropy of transformation distribution, truly reflect the uncertainty of registration?

\begin{figure}
	\centering
	\includegraphics[height=3.1cm]{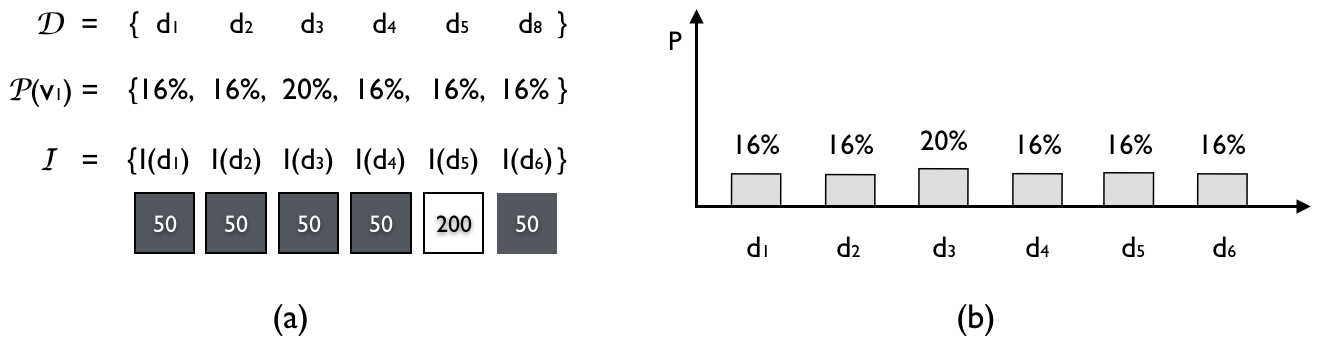}
	\vspace{-3mm}
	\caption{(a)The RWIR setting of a hypothetical example; (b)Bar chart of the transformation distribution $\mathcal{P}(v_1)$.}
	\vspace{-5mm}
	\label{fig:correct2}
\end{figure}

In a hypothetical RWIR example, assuming $v_1$ on $I_f$ is the voxel we want to register. As shown in Fig.2(a), $v_1$'s transformation space $\mathcal{D} = \{\mathbf{d}_k\}_{k=1}^6$ is a set of 6 displacement vectors. $\mathcal{P}(v_1)=\{P_k(v_1)\}_{k=1}^6$ is the computed distribution of $\mathcal{D}$. The corresponding labels for displacement vectors in $\mathcal{D}$ are image intensities stored in $\mathcal{I}=\{I(\mathbf{d_k})\}_{k=1}^6$. For clarity, suppose that there are only two different intensity values in $\mathcal{I}$, one is 50 and the other is 200. The color of squares in Fig.2(a) indicates the appearance of that intensity value. We can observe that $\mathbf{d_3}$ has the highest probability in $\mathcal{D}$, hence its corresponding intensity $I(\mathbf{d_3})=50$ will be assign to the registered $v_1$. 

Fig.2(b) is a bar chart illustrating the transformation distribution $\mathcal{P}(v_1)$. Although $\mathcal{P}(v_1)$ has its mode at $P_3(v_1)$ , the whole distribution is more or less uniformly distributed. The transformation distribution's entropy $E(\mathcal{P}(v_1))\approx2.58$ is close to the maximal. Therefore, the conventional uncertainty measure will suggest that the registration uncertainty of $v_1$ is high. Once clinicians knew its high amount of registration uncertainty, they would place less confidence in $v_1$'s current appearance.

The conventional way to quantify the registration uncertainty seems useful. However, its correctness is questionable. In the same $v_1$ RWIR example, let's take into account the intensity value $I(\mathbf{d_k})$ associated with each $\mathbf{d_k}$ and form an intensity distribution. As shown in Fig.3(a), even if $\mathbf{d_1},\mathbf{d_2},\mathbf{d_4}$ and $\mathbf{d_6}$ are different displacement vectors, they correspond to the same intensity value as the most likely displacement vector $\mathbf{d_3}$. As we accumulate the probability for all intensity values in $\mathcal{I}$, it is clear that 50 is the dominate intensity. Interestingly, despite being suggested of having high registration uncertainty by the conventional uncertainty measure, the intensity distribution in Fig.3(b) indicates that the appearance of registered $v_1$ is quite trustworthy. In addition, the entropy of the intensity distribution is as low as 0.63, which also differs from the high entropy value computed from the transformation distribution.

\begin{figure}
	\centering
	\includegraphics[height=3.2cm]{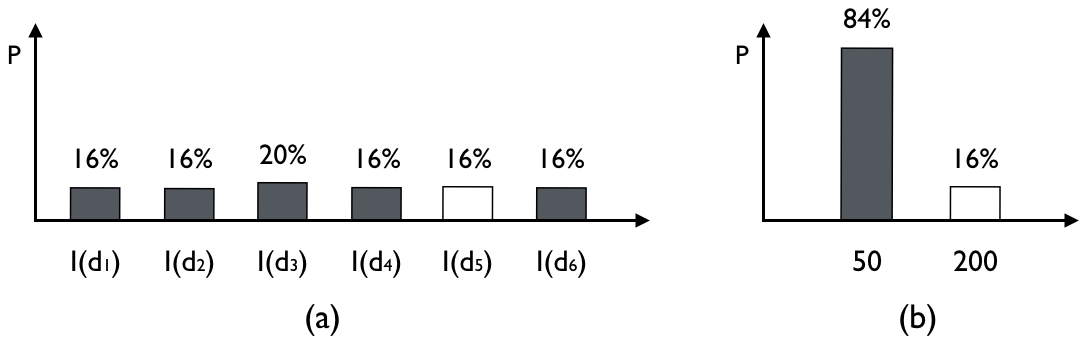}
	\vspace{-3mm}
	\caption{(a)Bar chart of the transformation distribution $\mathcal{P}(v_1)$ taking into account $I(\mathbf{d_k})$. The color of each bar indicates the appearance of $I(\mathbf{d_k})$; (b)Intensity distribution of the registered $v_1$.}
	\vspace{-5mm}
	\label{fig:correct2}
\end{figure}

This counter-intuitive example implies that high transformation uncertainty does not guarantee high registration uncertainty. In fact, the amount of transformation uncertainty can hardly guarantee any useful information about the registration uncertainty at all. More precisely, in the PIR setting, the transformation $R_T$ is modeled as a random variable. The corresponding label $R_L$, consisting of intensity values or tissue classes, is a function of $R_T$, so it is also a random variable. Even if $R_T$ and $R_L$ are intuitively correlated, given different hyper parameters and priors, there is no guaranteed statistical correlation between these two random variables. Therefore, it's inappropriate to measure the statistics of $R_L$ by the summary statistics of $R_T$.

In practice, for many PIR approaches, the likelihood term is often based on voxel intensity differences. In case there is no strong informative prior, these approaches tend to estimate ``flat" transformation distribution for voxels in homogeneous intensity regions. Transformation distributions of these voxels are usually more diverse than their intensity distributions, and therefore they are typical examples of how the conventional uncertainty measure, that is using the transformation uncertianty to quantify the registration uncertainty, tends to report false results \cite{Simpson,Lotfi}. 

In the following real data example, as shown in Fig.4(a), $I_f$ and $I_m$ are two brain MRI images arbitrarily chosen from the CUMC12 dataset. After performing RWIR, we obtain the registered moving image $I_{rm}$. To give more insight into the misleading defect of conventional uncertainty measures, we take a closer look at two voxels, $v_c$ at the center of a white matter area on the zoomed $I_{rm}$, and $v_e$ near the boundary of a ventricle. As can be seen from Fig.4(b), the transformation distribution of $v_c$ is more uniformly distributed than that of $v_e$. Therefore, conventional entropy-based methods will report $v_c$ having higher registration uncertainty than $v_e$. However, like the hypothetical example in Fig.3, we take into account the corresponding intensities and form a new intensity distribution. Since the intensity distribution is no longer categorical, we can employ other summary statistics, such as the variance, to measure the uncertainty. It turns out that the registered $v_e$ has larger intensity variance than $v_c$, which again reveals that the conventional uncertainty measure is misleading. 

\begin{figure}[t]
	\centering
	\includegraphics[height=5.3cm]{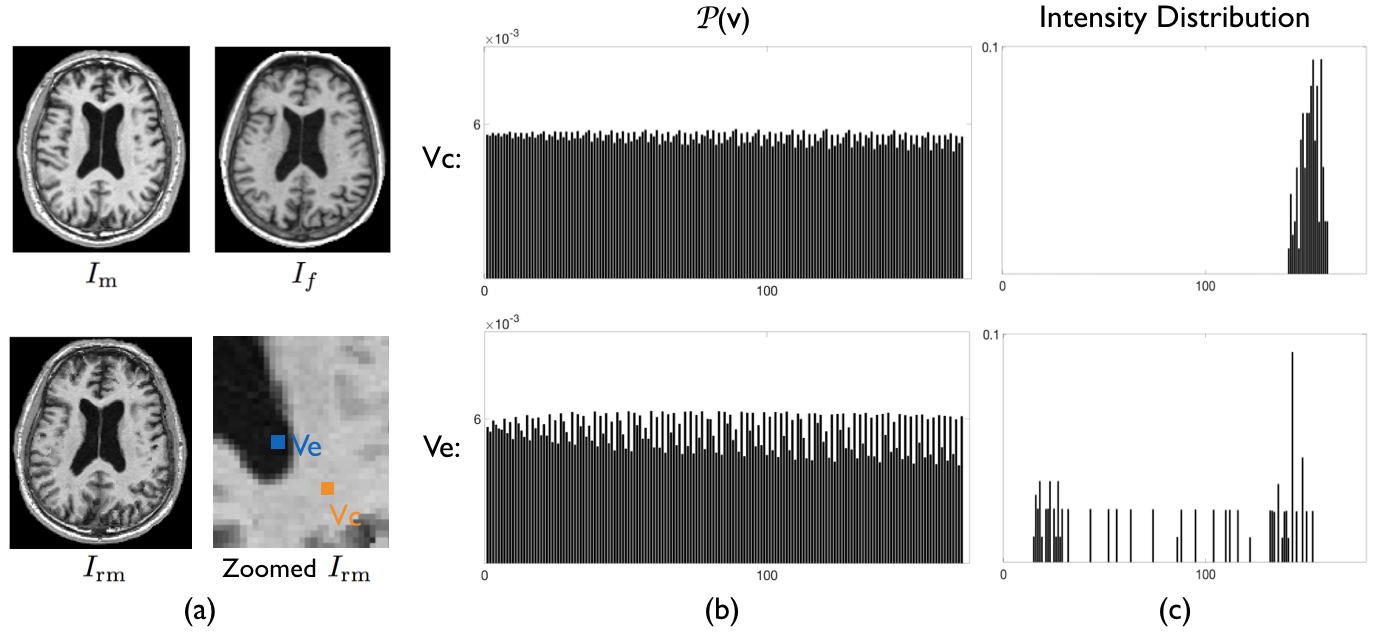}
	\vspace{-4mm}
	\caption{(a)Input and result of the CUMC12 data example; (b)The transformation distribution of $v_c$ and $v_e$ in the RWIR; (c)Intensity distributions of registered $v_c$ and $v_e$.}
	\vspace{-6mm}
	\label{fig:correct2}
\end{figure}

\section{Important yet Subtle Issues in PIR}

Point-estimate registration methods output a unique transformation, and establish the correspondence $I_\mathrm{rm}$ by assigning the corresponding label of its transformation to each voxel on $I_\mathrm{f}$. PIR methods output a transformation distribution, yet they still seek to establish a ``point-estimate" correspondence. Since the transformation mode is the most likely transformation, the common standard for PIR to establish the correspondence $I_\mathrm{rm}$ is assigning the corresponding label of its transformation mode to each voxel on $I_\mathrm{f}$. However, is it reasonable to determine the correspondence solely by the transformation mode?

\begin{figure}[t]
	\centering
	\includegraphics[height=3.0cm]{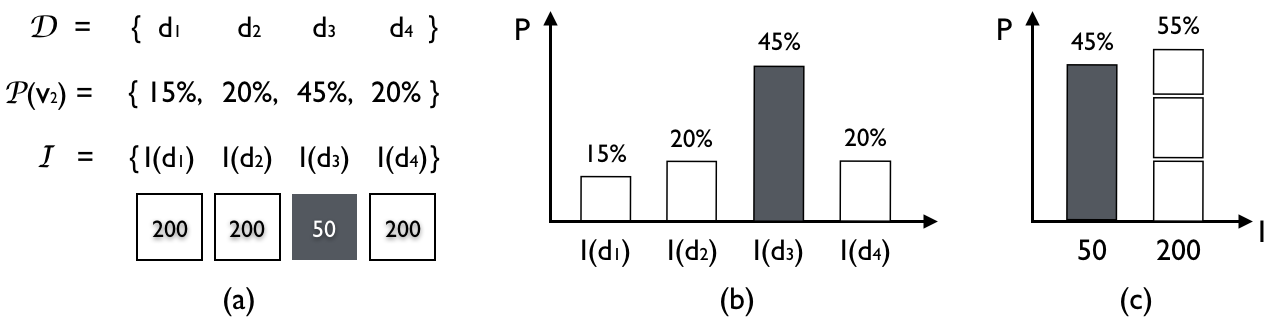}
	\vspace{-3mm}
	\caption{(a)The RWIR setting of the second hypothetical example; (b)Bar chart of the transformation distribution $\mathcal{P}(v_2)$ taking into account $I(\mathbf{d_k})$; (c)Intensity distribution of the registered $v_2$.}
	\vspace{-5mm}
	\label{fig:correct2}
\end{figure}

In another hypothetical example, assuming $v_2$ on $I_f$ is the voxel we want to register. As shown in Fig.5(a), the transformation $\mathcal{D} = \{\mathbf{d}_k\}_{k=1}^4$ is a set of 4 displacement vectors. $\mathcal{P}(v_2)=\{P_k(v_2)\}_{k=1}^4$ is the estimated distribution of $\mathcal{D}$. The corresponding intensity labels of all displacement vectors in $\mathcal{D}$ are stored in $\mathcal{I}=\{I(\mathbf{d_k})\}_{k=1}^4$. In RWIR, the transformation mode $\mathbf{d_m}$ is the displacement vector with the highest probability. Therefore, $\mathbf{d}_3$ is the transformation mode, and $I(\mathbf{d_m})=I(\mathbf{d_3})$ will be assigned to the registered $v_2$. The probability of $\mathbf{d}_3$ is considerably higher than that of other displacement vectors. Based on the relatively low entropy of the transformation distribution $\mathcal{P}(v_2)$, the intensity of registered $v_2$ should be trustworthy. However, once again we take into account the intensity value $I(\mathbf{d_k})$ associated with each $\mathbf{d_k}$, and form an intensity distribution. Surprisingly enough, Fig.5(c) shows that the corresponding intensity of the transformation mode $I(\mathbf{d_m})=50$ is no longer the most likely intensity. Displacement vectors $\mathbf{d_1},\mathbf{d_2}$ and $\mathbf{d_4}$ are all less likely transformations, yet their combined corresponding intensities outweigh $I(\mathbf{d}_3)$. 
 
The above example implies that the corresponding label of the transformation mode can differ from the most likely correspondence that is given by the full transformation distribution. This example makes sense because in the previous section we have pointed out that, in PIR, the transformation $R_T$ and correspondence $R_L$ are both regarded as random variables. Since there is no guaranteed statistical correlation between $R_T$ and $R_L$, the mode of $R_T$'s distribution is not guaranteed to be the mode of $R_L$'s distribution.

\begin{figure}[t]
	\centering
	\includegraphics[height=5.5cm]{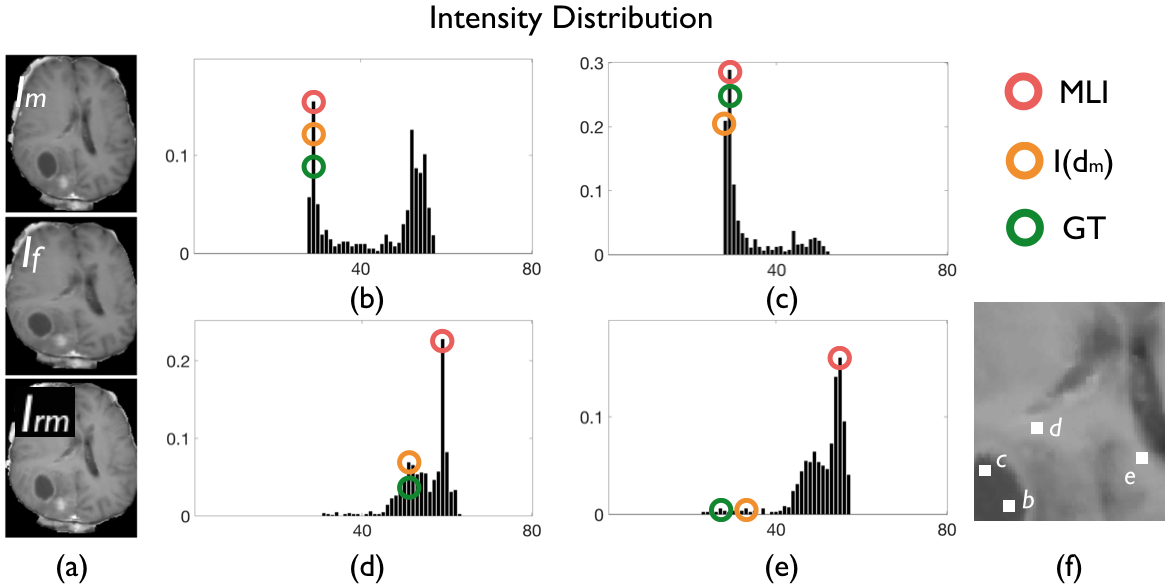}
	\vspace{-3mm}
	\caption{(a)Input and result of the BRATS data example; (b,c,d,e)Intensity distributions of $v_b$,$v_c$,$v_d$ and $v_e$; (f)Approximate locations of $v_b$,$v_c$,$v_d$ and $v_e$.}
	\vspace{-5mm}
	\label{fig:useful}
\end{figure}

As illustrated in Fig.6(a), we generate another example that register a MRI image $I_f$, which is arbitrarily chosen from the BRATS dataset, with synthetically distorted itself using RWIR. In this example, we investigate intensity distributions of four registered voxels $v_b,v_c,v_d$ and $v_e$, which are shown in Fig.6(b),(c),(d),(e) respectively. In Fig.6, the red circle indicates the Most Likely Intensity (MLI) given by the full transformation distribution, the orange circle indicates the corresponding intensity of the transformation mode $I(\mathbf{d_m})$, and the green circle is the Ground Truth (GT) intensity. We can observe that for $v_b$, the MLI and $I(\mathbf{d_m})$ are both equal to the GT. On the other hand, for $v_c, v_d$ and $v_e$, their MLIs are indeed not equal to their $I(\mathbf{d_m})$. This experiment does support our point of view that the corresponding label of the transformation mode $I(\mathbf{d_m})$ is not guaranteed to be the most likely label given by the full transformation distribution. However, at this stage, we can not conclude which one is better with respect to the registration accuracy for PIR.

As we conduct more experiments, we come across another interesting finding. As can be seen in Fig.6(c), the MLI of registered $v_c$ is equal to the GT intensity and more accurate than $I(\mathbf{d_m})$. Yet for $v_d$ and $v_e$, unexpectedly, it is their $I(\mathbf{d_m})$ more closer to the GT than their MLI. Voxels like $v_d$ and $v_e$ can be found very frequently in our experiments using other real data. This surprising result indicates that utilizing the full transformation distribution can actually give worse estimation than using the transformation mode alone.

Some existing researches have reported that it was beneficial to utilize the registration uncertainty, which is information obtained from the full transformation distribution, in some PIR-based tasks \cite{Heinrich,Risholm2,Simpson3}. However, the above finding make us wonder whether utilizing the full transformation distribution could always improve the performance.
 
It is noteworthy that the above finding is based on RWIR. In PIR, the correlation between the transformation $R_T$ and the correspondence $R_L$ is influenced by the choice of hyper parameters and priors. Other PIR approaches that use different transformation, regularization and optimization models, hence having different hyper parameters and priors, can certainly yield different findings than RWIR. However, we still suggest that researchers should analyze and investigate the credibility of the full transformation distribution before using it.

 \vspace{-3mm}
\section{Summary}

Previous studies don't differentiate the transformation uncertainty from the registration uncertainty. In this paper, we point out that, in PIR the transformation $R_T$ and the correspondence $R_L$ are both random variables, so it is inappropriate to quantify the uncertainty of $R_L$ by the summary statistics of $R_T$. We have also raised attention to an important yet subtle aspect of PIR, that is whether it is reasonable to determine the correspondence of a registered voxel solely by the mode of its transformation distribution. We reveal that the corresponding label of the transformation mode is not guaranteed to be the most likely correspondence given by the full transformation distribution. Finally, we share our concerns with respect to another intriguing finding, that is utilizing the full transformation distribution can actually give worse estimation.

Findings presented in this paper are significant for the development of PIR. We feel it is necessary to share our findings to the registration community.

\end{document}